\pdfoutput=1
\documentclass[11pt]{article}

\usepackage[final]{acl}

\usepackage{times}
\usepackage{latexsym}
\usepackage{booktabs}
\usepackage{amsmath}
\usepackage{multirow}
\usepackage{xcolor}
\usepackage{arydshln}
\usepackage{float}

\usepackage[T1]{fontenc}
\usepackage[utf8]{inputenc}

\usepackage{microtype}

\usepackage{inconsolata}

\usepackage{graphicx}

\title{Comparing human and LLM proofreading in L2 writing: \\ Impact on lexical and syntactic features}

\author{
Hakyung Sung\textsuperscript{1}\quad 
Karla Csuros\textsuperscript{2,1}\thanks{Contributed equally to the study.}\quad 
Min-Chang Sung\textsuperscript{3,1}\footnotemark[1] \\
\textsuperscript{1}LCR-ADS Lab, Linguistics, University of Oregon \\
\textsuperscript{2}West University of Timisoara \\
\textsuperscript{3}Gyeongin National University of Education \\ %add department?
}

\begin{document}
\maketitle
\begin{abstract}
This study examines the lexical and syntactic interventions of human and LLM proofreading aimed at improving overall intelligibility in identical second language writings, and evaluates the consistency of outcomes across three LLMs (ChatGPT-4o, Llama3.1-8b, Deepseek-r1-8b). Findings show that both human and LLM proofreading enhance bigram lexical features, which may contribute to better coherence and contextual connectedness between adjacent words. However, LLM proofreading exhibits a more generative approach, extensively reworking vocabulary and sentence structures, such as employing more diverse and sophisticated vocabulary and incorporating a greater number of adjective modifiers in noun phrases. The proofreading outcomes are highly consistent in major lexical and syntactic features across the three models.
\end{abstract}

\section{Introduction}
The use of generative large language models (LLMs) in second language (L2) writing has gained popularity for providing real-time feedback on vocabulary, grammar, and style (e.g., \citealp{han-etal-2024-llm, meyer2024using}). These models offer immediate corrective suggestions, enhancing the precision and quality of L2 writing---a role once largely filled by human editors with expertise. As LLMs increasingly replace or supplement human intervention, questions arise about their impact on L2 writings.

While previous studies have concentrated on general error correction through LLM proofreading (e.g., \citealp{heintz2022comparing, su2023collaborating, wu2023chatgpt, katinskaia2024gpt}), recent studies have shown that LLMs do not consistently outperform state-of-the-art supervised grammatical error correction models on minimal-edit benchmarks, often producing more fluency-oriented rewrites instead \citep{davis2024prompting}. This tendency stems in part from the fact that LLMs, by default, generate transformative fluency corrections rather than minimal edits when processing ungrammatical text (e.g., \citealp{coyne2023analyzing, fang2023chatgpt, loem2023exploring}). However, little research has examined how this generative rewriting behavior affects broader lexical and syntactic characteristics of L2 writing compared to human proofreading, especially when the proofreading goal extends beyond grammatical accuracy to overall intelligibility. Moreover, it remains unclear whether different LLMs yield consistent proofreading outcomes. This study addresses these gaps by posing three guiding questions: (1) What are the similarities and differences in lexical features between human proofreading and LLM proofreading of L2 writings? (2) What are the similarities and differences in syntactic features between human proofreading and LLM proofreading of L2 writings? (3) Do three different LLMs provide consistent proofreading outcomes in terms of lexical and syntactic features in L2 writing? 

Our findings show that while both human and LLM proofreading enhance lexical and syntactic features, LLMs 
are more likely to make more extensive lexical and syntactic edits. By quantifying these changes through a range of lexical and syntactic indices, we reveal that LLMs favor more generative rewrites, which may improve fluency but risk altering nuance or inflating perceived proficiency. 

\section{Background}

\subsection{Proofreading in L2 writing}

Proofreading is a complex issue in writing research, particularly for L2 writers, as it involves varying scopes of interventions. Traditional definitions of proofreading often restrict it to surface-level error correction that focuses on resolving orthographic and grammatical errors without altering content \cite{carduner2007teaching,hyatt2017student}. However, research shows that professional human proofreaders occasionally restructure content to improve the logical flow of ideas and make the writing easier to understand \cite{salter2019proofreading}. Noting these varying practices in proofreading, \citet[p. 167]{harwood2009proofreading} provided a quite general definition of proofreading as ``[any] third-party interventions (entailing written alteration) on assessed work in progress.''

Previous studies have shown that human proofreading displays variability not just in scope, but also in quality. \citet{harwood2018proofreaders} found that 14 proofreaders made between 113 and 472 changes to the same L2 learner essay, with some interventions improving clarity and others introducing new errors, leading to inconsistent quality. Similarly, \citet{shafto2015proofreading} argued that proofreading is a highly attention-dependent task, meaning that symptoms such as tiredness can heavily impact human proofreaders' ability to detect and correct ungrammatical and unnatural expressions.

The debate surrounding the adequacy of L2 proofreading is also characterized by varying perspectives from stakeholders (i.e., students, faculty, researchers). While L2 students often seek proofreading services to improve their grades or enhance their writing skills, some faculty view such assistance as a form of academic dishonesty \cite{salter2019proofreading, turner2011rewriting}. Despite these divergent opinions, there is a general consensus that proofreaders can significantly enhance language accuracy and clarity in L2 writing, provided that the original authorial voice is maintained \cite{turnerafterword,warschauer2023affordances,zou2024impact}.

\subsection{LLMs in L2 writing and proofreading}

While automated written corrective feedback has been present in L2 classrooms for over a decade (cf. \citealp{wilson2014does}), recent research is now exploring how LLM assistants can be incorporated into holistic writing workflows \cite{zhao2024impact}. Researchers examine the integration of the LLM in prewriting \cite{xiao2024chatgpt} and postwriting stages \cite{osawa2024integrating}, as well as its role in fostering metacognitive skills through iterative revisions that include editing and proofreading \cite{su2023collaborating,warschauer2023affordances,zou2024impact}.

Among these LLM integrations, several studies have highlighted the capabilities of LLM proofreading (or more broadly, editing). For instance, \citet{su2023collaborating} found that ChatGPT effectively assessed grammar, clarified meaning, and suggested lexical and syntactic refinements. Similarly, \citet{yan2024l2} observed that ChatGPT identified and corrected a range of linguistic errors—including lexical (e.g., word choice, idioms), grammatical (e.g., verb tense, articles), structural (e.g., run-on or fragmented sentences), mechanical (e.g., spelling, punctuation), and stylistic (e.g., formality) aspects.

Few studies have compared LLM proofreading directly to human revisions. For instance, \citet{heintz2022comparing} compared outputs edited by LLMs with those revised by human editors using sentences written by non-native English speakers. They found that while Wordvice AI\footnote{\url{https://wordvice.ai/proofreading}} achieved near-human accuracy (77\%) in correcting grammar and spelling errors, it lagged behind human editors in areas like vocabulary refinement and fluency adjustments. Similarly, \citet{jiang2023exploring} analyzed 2,197 T-units\footnote{A T-unit is often defined as the minimal grammatical unit, comprising a single independent clause plus any subordinate clauses or dependent phrases attached to it \cite{lu2010automatic}.} and 1,410 sentences from weekly writing samples of 41 Chinese students in an online high school language program at a U.S. university. They found that ChatGPT-4 achieved high precision (88\%) in correcting errors at the T-unit level (in comparison to human judgments), but sometimes overcorrected valid sentences or misinterpreted context-dependent issues, such as ambiguous word order and culturally embedded idioms. 

\subsection{Summary of findings and research gaps}
To briefly summarize, previous research has demonstrated that proofreading in L2 writing is highly variable in both scope and quality, with interventions ranging from surface-level corrections to content restructuring. Recently, LLMs have been shown to offer performance comparable to, or even surpassing, that of human editors in L2 writing proofreading, although they exhibit limitations in context-sensitive judgment and cultural awareness.

Despite these insights, still little is known about the fine-grained linguistic interventions that could be made by LLMs compared to human proofreaders. Additionally, existing research has focused primarily on grammatical error detection and correction, overlooking broader language use. For example, although LLMs may facilitate vocabulary expansion, it remains unclear how their suggestions differ from those of human proofreaders, and detailed syntactic changes remain underexplored. Moreover, most studies have examined only one type of LLM, leaving open the question of whether these linguistic changes are specific to one model or generalizable across other LLMs.

\section{Methods}

\subsection{Dataset}
This study utilizes the ICNALE Edited Essays dataset, one of the publicly available corpora within the International Corpus Network of Asian Learners of English (ICNALE) project (\citealp{Ishikawa2018TheIE, ishikawa2021asian}). The dataset comprises 656 essays written by 328 L2 learners and their edited versions produced by professional native English-speaking proofreaders.

The L2 participants were college students learning English in ten regional contexts: Japan (JPN), Korea (KOR), China (CHN), Taiwan (TWN), Indonesia (IDN), Thailand (THA), Hong Kong (HKG), the Philippines (PHL), Pakistan (PAK), and Singapore (SIN). Each participant wrote two argumentative essays in response to the prompts: (1) ``It is important for college students to have a part-time job'' and (2) ``Smoking should be completely banned at all restaurants''.

\subsubsection{Rationale for dataset selection and representativeness}
The ICNALE dataset was chosen for three main reasons. First, it provides paired original and professionally proofread versions, allowing for direct comparison with LLM-generated outputs. Second, it includes explicit L2 proficiency labels, facilitating stratified analyses across proficiency levels. Last, it offers balanced regional coverage across ten Asian countries or regions (see Table~\ref{tab:1}). However, we acknowledge that broad generalizations to other genres or demographic groups (e.g., narrative writing, younger learners) must be made with caution.

\subsubsection{Proficiency band}
All participants were classified into four L2 proficiency bands (linked to the Common European Framework of Reference for Languages) based on their recent scores in standardized English tests (e.g., TOEFL, TOEIC) or their performance in a standard receptive vocabulary test\footnote{The vocabulary test consists of 50 multiple-choice items designed to measure vocabulary knowledge within the 1,000--5,000 word range. A typical item (from the 4,000-word level) presents a short sentence containing a target word and asks test-takers to select the most appropriate definition.} \citep{nation2007vocabulary}.  Table \ref{tab:1} shows the proficiency distribution of each regional learner group.

\begin{table}[ht]
\centering
\resizebox{0.8\linewidth}{!}{%
\begin{tabular}{lcccccc}
\toprule
Region & A2\_0 & B1\_1 & B1\_2 & B2\_0 & Total \\
\midrule
JPN  & 10 & 10 & 10 & 10 & 40 \\
KOR  & 10 & 10 & 10 & 10 & 40 \\
CHN  & 10 & 10 & 10 & 10 & 40 \\
TWN  & 10 & 10 & 10 & 10 & 40 \\
IDN  & 10 & 10 & 10 & 3 & 33 \\
THA  & 10 & 10 & 10 & 2 & 32 \\
HKG  & -- & 10 & 10 & 10 & 30 \\
PHL  & -- & 10 & 10 & 10 & 30 \\
PAK  & -- & 10 & 10 & 3 & 23 \\
SIN  & -- & -- & 10 & 10 & 20 \\
\midrule
Total & 60 & 90 & 100 & 78 & 328 \\
\bottomrule
\end{tabular}}
\caption{Distribution of participants by region and proficiency}
\label{tab:1}
\end{table}

\subsubsection{Proofreading process and proofreader profiles}
The ICNALE project recruited five experienced proofreaders with strong academic backgrounds and extensive experience in editing scholarly work. Their profiles are summarized in Table~\ref{tab:2}.

\begin{table}[ht]
\centering
\resizebox{\linewidth}{!}{%
\begin{tabular}{lccccc}
\toprule
ID & Age & Sex & Degree & Experience (years) & L1 English \\
\midrule
A & 28 & Female & BA & 3  & Canadian \\
B & 32 & Female & MS & 5  & Australian \\
C & 27 & Female & BS & 3  & American \\
D & 38 & Female & BS & 10 & British \\
E & 31 & Female & PhD & 2  & Australian \\
\bottomrule
\end{tabular}}
\caption{Profiles of proofreaders in the ICNALE project}
\label{tab:2}
\end{table}

As documented in the ICNALE project, the professional proofreaders were tasked with editing errors and inappropriate wording to ensure that each essay became fully intelligible (\citealp{ishikawa2021asian}, p. 496). No standardized rubric or adjudication mechanism was imposed at the original corpus compilation stage. All revisions were performed in MS Word using the Track Changes function, which allowed every edit, addition, or deletion to be recorded.

\begin{figure*}[!]
    \centering
    \includegraphics[width=0.75\textwidth]{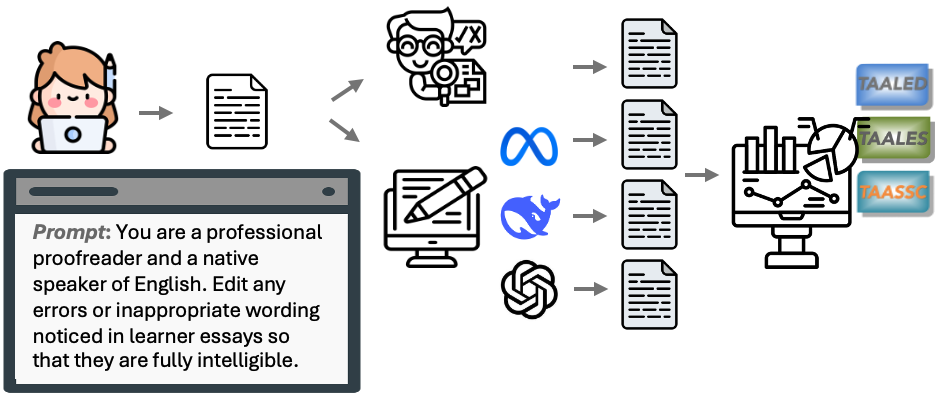}
    \caption{Overview of the experiment}
    \label{fig:1}
\end{figure*}

A calibration study in which all five proofreaders revised the same eight essays revealed substantial variability in editing behavior (cf. \citealp{Ishikawa2018TheIE}, p. 122). The number of edited word tokens ranged from 40.00 to 59.63—a difference of 19.63 tokens, or 40.97\% of the average. \citet{ishikawa2021asian} attributed this variation to the inherent subjectivity of human editing, shaped by individual judgments of intelligibility.

\subsection{LLM selection and prompt design}
Figure \ref{fig:1} outlines the experiment. First, to compare the human proofreading in the ICNALE project with LLM proofreading, we selected three text-generating LLMs: GPT-4o (used in ChatGPT, accessed via OpenAI's API; \citealp{achiam2023gpt}, hence we called them \textit{Chatgpt-4o}), \textit{Llama3.1-8b} \cite{touvron2023llama}, and \textit{Deepseek-r1-8b} \cite{guo2025deepseek}. ChatGPT-4o was chosen due to its widespread accessibility, although its underlying parameter count and architecture remain proprietary. In contrast, both Llama3.1-8b and Deepseek-r1-8b are open models with 8 billion parameters that are lightweight enough for local installations, with Deepseek-r1-8b being a distilled version of Llama3.1-8b.

Each model was tasked with reading the original L2 writings and generating a proofread version based solely on a standardized prompt, with no access to additional learner information. The exact prompt used was as follows: ``You are a professional proofreader and a native speaker of English. Edit any errors or inappropriate wording noticed in learner essays so that they are fully intelligible. Return only the final edited version of the essay. Do not include any explanations, comments, reasoning, or additional thoughts in your response.'' This prompt was designed to align with the instructions given to ICNALE proofreaders—``They were asked to edit any error or inappropriate wording noticed in learner essays so that they could be fully intelligible. They were also required not to ‘rewrite’ the original texts, that is, not to add new content or to alter organization'' (\citealp{ishikawa2021asian}, p. 496)—ensuring consistency with the human proofreading protocol for fair comparison.

\subsection{Lexical and syntactic analyses}
The proofread-and-generated texts, along with the learner and edited texts in the ICNALE dataset, were processed to extract lexical and syntactic features using the source codes of publicly available NLP tools: TAALED (cf. \citealp{kyle2024evaluating}), TAALES (cf. \citealp{kyle2018tool}) and TAASSC (cf. \citealp{kyle2018measuring}). 
We measured lexical and syntactic aspects of the learner and proofread essays based on the concept of linguistic complexity, which provides a descriptive-analytic framework for L2 production \cite{bulte2012defining,bulte2024complexity}.

\subsubsection{Lexical features}

Lexical features were evaluated in terms of two aspects: diversity and sophistication. Lexical diversity indices reflect vocabulary variation and repetition, with higher scores indicating a broader vocabulary range and fewer repetitions. In this study, we employ common measures such as the number of unique words and the moving-average type-token ratio—the latter mitigating the impact of text length on traditional lexical diversity measures \cite{kyle2024evaluating}.

Lexical sophistication indices, on the other hand, focus on measuring the use of advanced words \cite{laufer1995vocabulary, meara2001p}. They are typically assessed based on relative word frequency, semantic concreteness, and domain or register distinctiveness, with less frequent, less concrete, and more domain-specific words generally considered more sophisticated \cite{kyle2018tool}. We also incorporate the concept of ngram sophistication by analyzing associations and dependency relations within bigrams \cite{kyle2021automatically}.

\subsubsection{Syntactic features}

Syntactic features can be examined from multiple perspectives. Traditional approaches, such as measuring the average length of T-units, focus on the overall length of syntactic structures and operate under the assumption that longer units generally indicate greater complexity \cite{lu2010automatic, lu2011corpus}.

In contrast, fine-grained syntactic complexity indices \cite{kyle2018measuring} provide a more nuanced analysis by capturing specific structural characteristics rather than relying on surface-level measures like sentence length. These indices are often categorized into clausal-level (e.g., nominal subjects per clause), phrasal-level (e.g., dependents per nominal, including adjectives and prepositions), and morphosyntactic-level features (e.g., use of past tense).

To the best of our knowledge, there is no consensus on which fine-grained indices reliably capture syntactic complexity as perceived by human judges. Nevertheless, L2 writing studies suggest that higher-proficiency learners (identified by human ratings) tend to use more elaborated noun phrases (e.g., \citealp{biber2011should}).

\subsection{Statistical methods}

\subsubsection{Evaluating linguistic features across groups}

Prior to statistical analyses, we confirmed that the five groups of texts (i.e., original [ORIG], human-proofread [EDIT], and the three LLM-proofread versions) were largely comparable in length.\footnote{The differences in the number of word tokens relative to the original text were: EDIT: –1.02, ChatGPT-4o: +6.13, Llama3.1-8b: –3.38, and Deepseek-r1-8b: –15.11\textsuperscript{***}.} This comparability, with the exception of Deepseek-r1-8b, indicates that subsequent improvements in lexical and syntactic domains are not simply due to different text lengths.

\begin{table*}[!htbp]
\centering
\resizebox{\textwidth}{!}{%
\begin{tabular}{lcccc}
\toprule
\textbf{Index} 
 & \textbf{EDIT }
 & \textbf{ChatGPT-4o} 
 & \textbf{Llama3.1-8b}
 & \textbf{Deepseek-r1-8b} \\
\midrule

\texttt{raw\_bg\_MI}
  & \textcolor{blue}{+0.35} / 1.80\textsuperscript{***} 
  & \textcolor{blue}{+0.65} / 3.30\textsuperscript{***} 
  & \textcolor{blue}{+0.62} / 3.17\textsuperscript{***} 
  & \textcolor{blue}{+0.60} / 3.03\textsuperscript{***} \\

\texttt{usf}
  & \textcolor{red}{-1.37} / 0.15
  & \textcolor{red}{-9.21} / 0.99\textsuperscript{***} 
  & \textcolor{red}{-8.48} / 0.91\textsuperscript{***} 
  & \textcolor{red}{-12.09} / 1.30\textsuperscript{***} \\

\texttt{b\_concreteness}
  & \textcolor{black}{+0.00} / 0.02    
  & \textcolor{red}{-0.15} / 0.83\textsuperscript{***} 
  & \textcolor{red}{-0.12} / 0.67\textsuperscript{***} 
  & \textcolor{red}{-0.21} / 1.11\textsuperscript{***} \\
  
\texttt{cw\_lemma\_freq\_log}
  & \textcolor{red}{-0.02} / 0.03 
  & \textcolor{red}{-0.30} / 0.54\textsuperscript{***} 
  & \textcolor{red}{-0.26} / 0.47\textsuperscript{***} 
  & \textcolor{red}{-0.37} / 0.67\textsuperscript{***} \\

\texttt{mattr} 
  & \textcolor{blue}{+0.01} / 0.18    
  & \textcolor{blue}{+0.07} / 2.20\textsuperscript{***} 
  & \textcolor{blue}{+0.08} / 2.63\textsuperscript{***}  
  & \textcolor{blue}{+0.10} / 3.41\textsuperscript{***} \\ 

\texttt{ntypes}
  & \textcolor{blue}{+0.63} / 0.05
  & \textcolor{blue}{+19.98} / 1.68\textsuperscript{***} 
  & \textcolor{blue}{+16.68} / 1.40\textsuperscript{***} 
  & \textcolor{blue}{+16.80} / 1.41\textsuperscript{***} \\
  
\bottomrule
\end{tabular}}
\caption{
Lexical features compared; For each index, two numbers are shown: the value on the \textit{left} indicates the unstandardized main effect coefficient, while the value on the \textit{right} (following the backslash) represents the standardized coefficient, calculated as the ratio of the coefficient to the residual standard deviation of the dependent variable; Significance vs.\ ORIG is marked ($\ast p<0.0125$, $\ast\ast p<0.0025$, $\ast\ast\ast p<0.00025$); negative values are red and positive values are blue; interaction effects are omitted.}
\label{tab:3}
\end{table*}

We calculated a range of 49 lexical and 143 syntactic indices from every text in the five groups and identified features showing significant between-group variance in two stages. First, we conducted visual inspection of box plots to exclude the indices with a great number of outliers, little individual variance, and/or unnoticeable mean differences. Second, we applied a linear mixed-effects model to each index, using Group (e.g., ORIG, EDIT, ChatGPT-4o) as a categorical fixed effect with ORIG as the baseline. Proficiency was included as a fixed effect that interacted with Group, and Participants were included as a random effect. We retained only those models that converged successfully to ensure reliable estimates. From these convergent models, we focused primarily on the main effect of the proofreading mode, while also examining whether any observed mode effects were moderated by Proficiency. These procedures yielded six lexical and nine syntactic indices. Detailed descriptions of each index are provided in Appendix \ref{apeA}. 

For each of these indices, we reported the results of four pairwise comparisons, between ORIG and human or LLM proofreading, from the linear mixed-effects models. To avoid a Type I error due to multiple comparisons, we applied a Bonferroni adjustment to the alpha level, reducing it from .05 to .0125.

\subsubsection{Evaluating consistency across LLMs}

The linear mixed-effects analyses informed us that the cross-model evaluation should exclude five more syntactic features, which showed multicollinearity or overlapping metrics. For the rest ten features,\footnote{Lexical features: \texttt{mattr}, \texttt{b\_concreteness}, \texttt{mcd}, \texttt{usf}, \texttt{cw\_lemma\_freq\_log}, and \texttt{raw\_bg\_MI}; Syntactic features: \texttt{nonfinite\_prop}, \texttt{amod\_dep}, \texttt{nominalization}, and \texttt{be\_mv}.} we calculated the standardized z-scores so that each metric contributed equally to a composite measure of overall lexical and syntactic complexity.

Next, we restructured the data so that each row represented an essay and each column contained the composite score derived from the output of a different model, treating these composite scores as ``ratings” of the same essay. We then calculated the Pearson correlation coefficients between the ratings for every pair of models' proofread output and computed Cronbach’s alpha \cite{cronbach1951coefficient} across these scores to assess their overall consistency. All datasets and code used for this analysis are available in the supplementary repository: \url{https://osf.io/mhtpg/?view_only=13ce0959a80e4d498b6761aba197bc83}.

\section{Results}

\subsection{Lexical features}

\begin{figure}[!]
    \centering
    \includegraphics[width=1\linewidth]{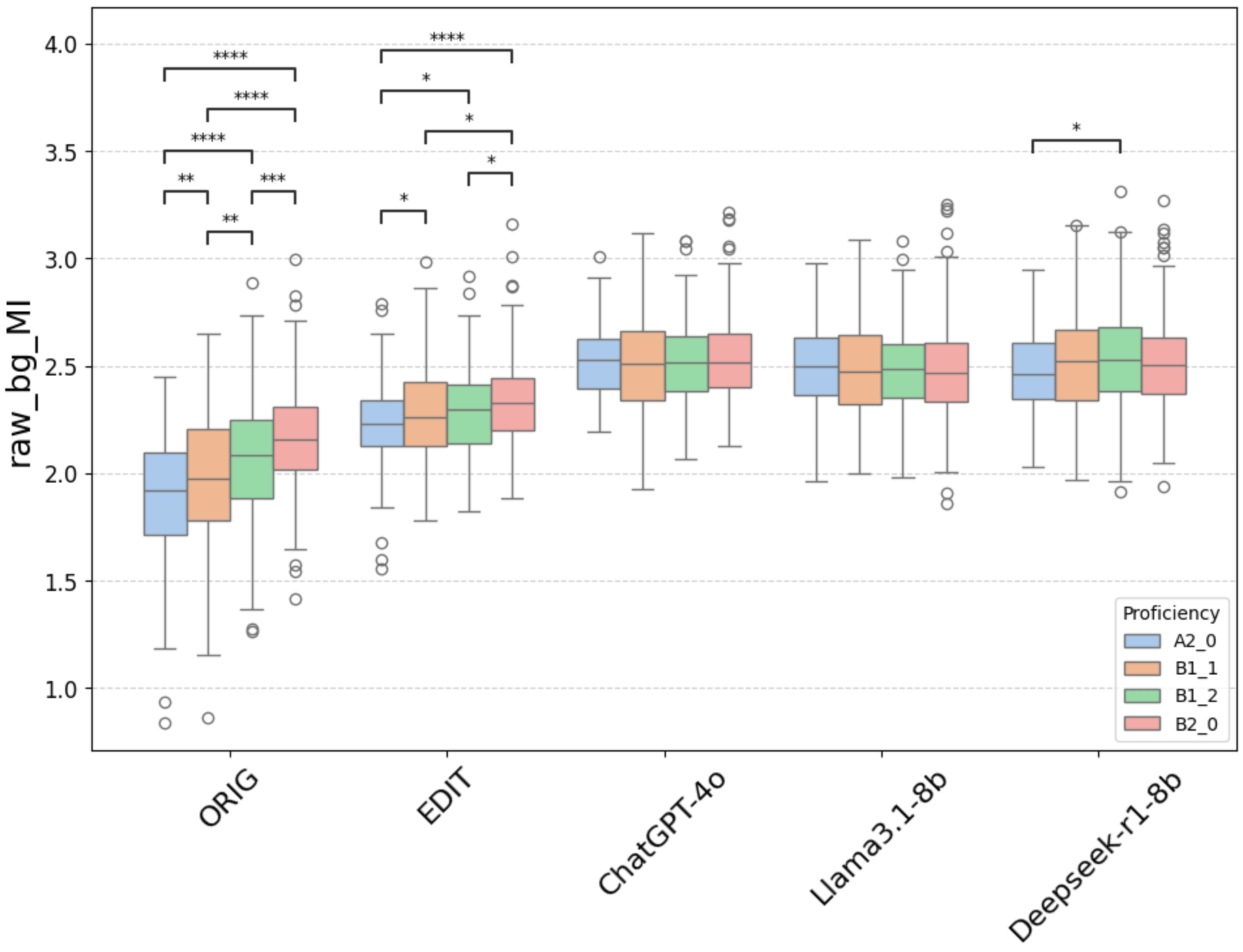}
    \caption{\texttt{raw\_bg\_MI} compared across ORIG, EDIT, and LLM-proofread texts by proficiency}
    \label{fig:2}
\end{figure}

Table \ref{tab:3} summarizes the analysis of the selected lexical sophistication and diversity features. First, all proofreading modes, including human editing, led to significantly higher bigram mutual information (\texttt{raw\_bg\_MI}) scores. This finding suggests that both human and LLM proofreading improved the lexical sophistication in terms of the coherence or contextual connectedness of adjacent words. However, LLM proofreading substantially increased \texttt{raw\_bg\_MI} to the extent that differences between lower and higher proficiency levels became less distinguishable (Figure \ref{fig:2}).

In contrast, only the LLM-proofread texts showed significant changes in additional lexical sophistication measures, including a shift toward more contextually distinctive words (\texttt{usf}), less concrete words (\texttt{b\_concreteness}), and lower-frequency content words (\texttt{cw\_lemma\_freq\_log}). Human proofreading, by comparison, did not produce significant differences in these measures.

\begin{figure}[!]
    \centering
    \includegraphics[width=1\linewidth]{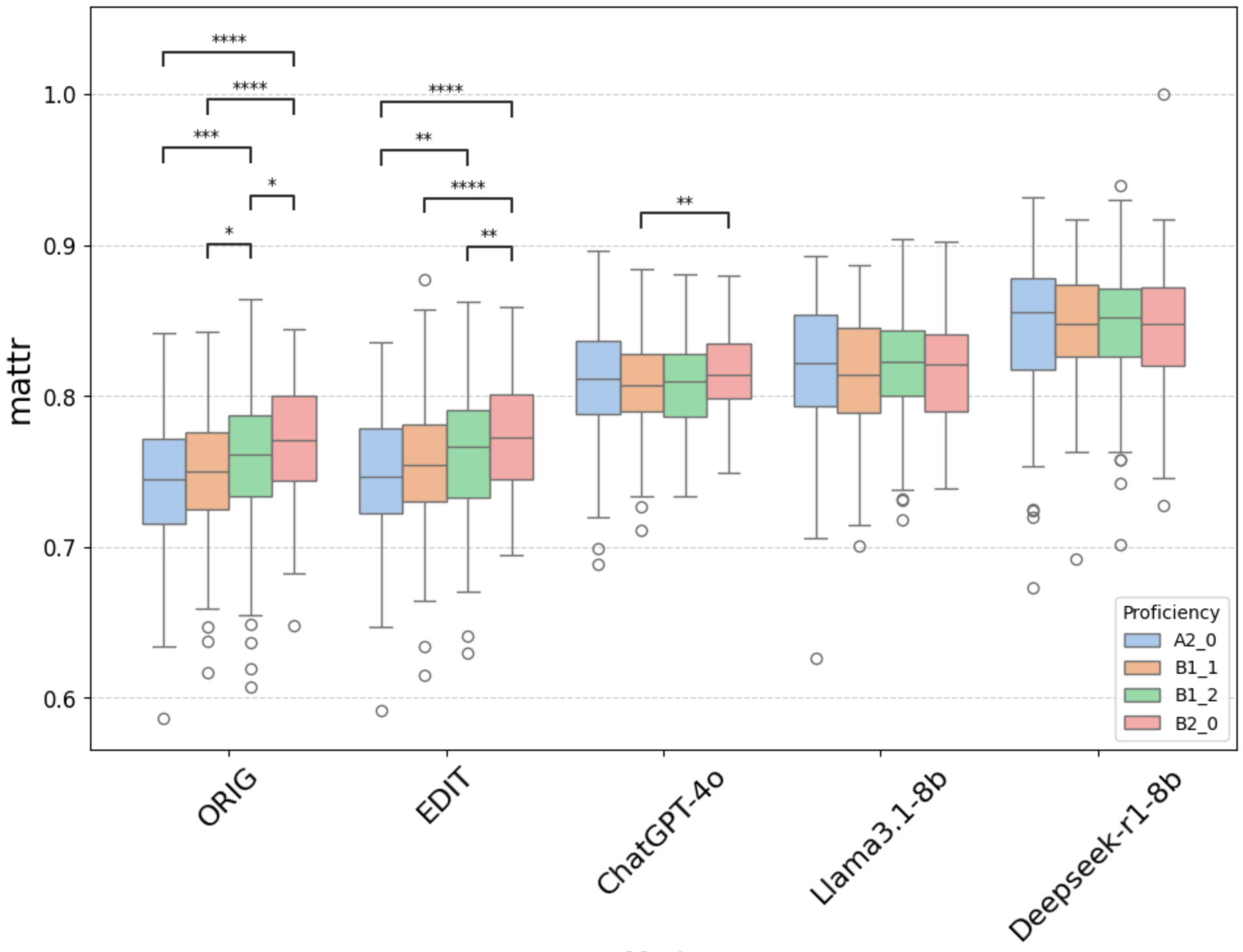}
    \caption{\texttt{mattr} compared across ORIG, EDIT, and LLM-proofread texts by proficiency}
    \label{fig:3}
\end{figure}

As for lexical diversity, significant improvements were observed only in the LLM-proofread texts, with increases in metrics such as \texttt{mattr} (Figure \ref{fig:3}) and \texttt{ntypes}, indicating a broader range of vocabulary use.

\begin{table*}[!]
\centering
\resizebox{\textwidth}{!}{%
\begin{tabular}{lcccc}
\toprule
\textbf{Index} 
 & \textbf{EDIT} 
 & \textbf{ChatGPT-4o} 
 & \textbf{Llama3.1-8b}
 & \textbf{Deepseek-r1-8b} \\
\midrule
\texttt{mltu}
 & \textcolor{red}{-115.49} / 0.31
 & \textcolor{red}{-105.73} / 0.28
 & \textcolor{blue}{+44.26} / 0.12
 & \textcolor{blue}{+118.42} / 0.31 \\\hdashline

\texttt{all\_clauses}
 & \textcolor{blue}{+15.55} / 0.10
 & \textcolor{blue}{+133.76} / 0.84\textsuperscript{***}
 & \textcolor{blue}{+99.12} / 0.62\textsuperscript{***}
 & \textcolor{blue}{+179.00} / 1.12\textsuperscript{***} \\

\texttt{nonfinite\_prop}
 & \textcolor{red}{-1.33} / 0.29
 & \textcolor{blue}{+2.01} / 0.44\textsuperscript{***}
 & \textcolor{blue}{+2.63} / 0.57\textsuperscript{***}
 & \textcolor{blue}{+5.52} / 1.20\textsuperscript{***} \\ \hdashline

\texttt{np}
 & \textcolor{red}{-21.30} / 0.08
 & \textcolor{blue}{+91.96} / 0.36\textsuperscript{**}
 & \textcolor{blue}{+41.27} / 0.16
 & \textcolor{blue}{+194.91} / 0.76\textsuperscript{***} \\

\texttt{np\_deps}
 & \textcolor{red}{-35.03} / 0.08
 & \textcolor{blue}{+79.21} / 0.17
 & \textcolor{blue}{+91.91} / 0.20
 & \textcolor{blue}{+217.81} / 0.47\textsuperscript{**} \\
 
\texttt{amod\_dep}
 & \textcolor{blue}{+17.54} / 0.01\textsuperscript
 & \textcolor{blue}{+137.65} / 0.75\textsuperscript{***}
 & \textcolor{blue}{+127.44} / 0.70\textsuperscript{***}
 & \textcolor{blue}{+204.54} / 1.12\textsuperscript{***} \\\hdashline

\texttt{nominalization}
 & \textcolor{blue}{+58.12} / 0.40\textsuperscript{**}
 & \textcolor{blue}{+152.04} / 1.05\textsuperscript{***}
 & \textcolor{blue}{+102.85} / 0.71\textsuperscript{***}
 & \textcolor{blue}{+213.63} / 1.47\textsuperscript{***} \\
 
\texttt{be\_mv}
 & \textcolor{blue}{+10.37} / 0.12
 & \textcolor{red}{-56.53} / 0.63\textsuperscript{***}
 & \textcolor{red}{-41.60} / 0.47\textsuperscript{**}
 & \textcolor{red}{-84.02} / 0.94\textsuperscript{***} \\
 
\texttt{past\_tense}
 & \textcolor{red}{-15.80} / 0.29\textsuperscript
 & \textcolor{red}{-17.38} / 0.32\textsuperscript
 & \textcolor{red}{-17.77} / 0.32\textsuperscript
 & \textcolor{red}{-19.31} / 0.35\textsuperscript{**} \\

\bottomrule
\end{tabular}}
\caption{Syntactic features compared; Interpretation of the table follows the same conventions described in Table \ref{tab:3}}
\label{tab:4}
\end{table*}

\subsection{Syntactic features}

Table \ref{tab:4} summarizes the analysis of the selected syntactic features. Regarding the mean length of T-units (\texttt{mltu}), neither human nor LLM proofreading produced a consistent pattern: human proofreading (EDIT) and ChatGPT-4o tended to reduce T-unit length, while Llama3.1-8b and Deepseek-r1-8b tended to increase it, suggesting no uniform effect on the length of minimal grammatical units.

At the clause level, all LLM-proofread texts showed a significant increase in the total number of clauses (\texttt{all\_clauses}) compared to the original learner essays, with Deepseek-r1-8b exhibiting the largest effect. Moreover, LLM-proofread texts contained a higher proportion of nonfinite clauses (\texttt{nonfinite\_prop}), whereas human editing resulted in a slight reduction in this index.

\begin{figure}
    \centering
    \includegraphics[width=1\linewidth]{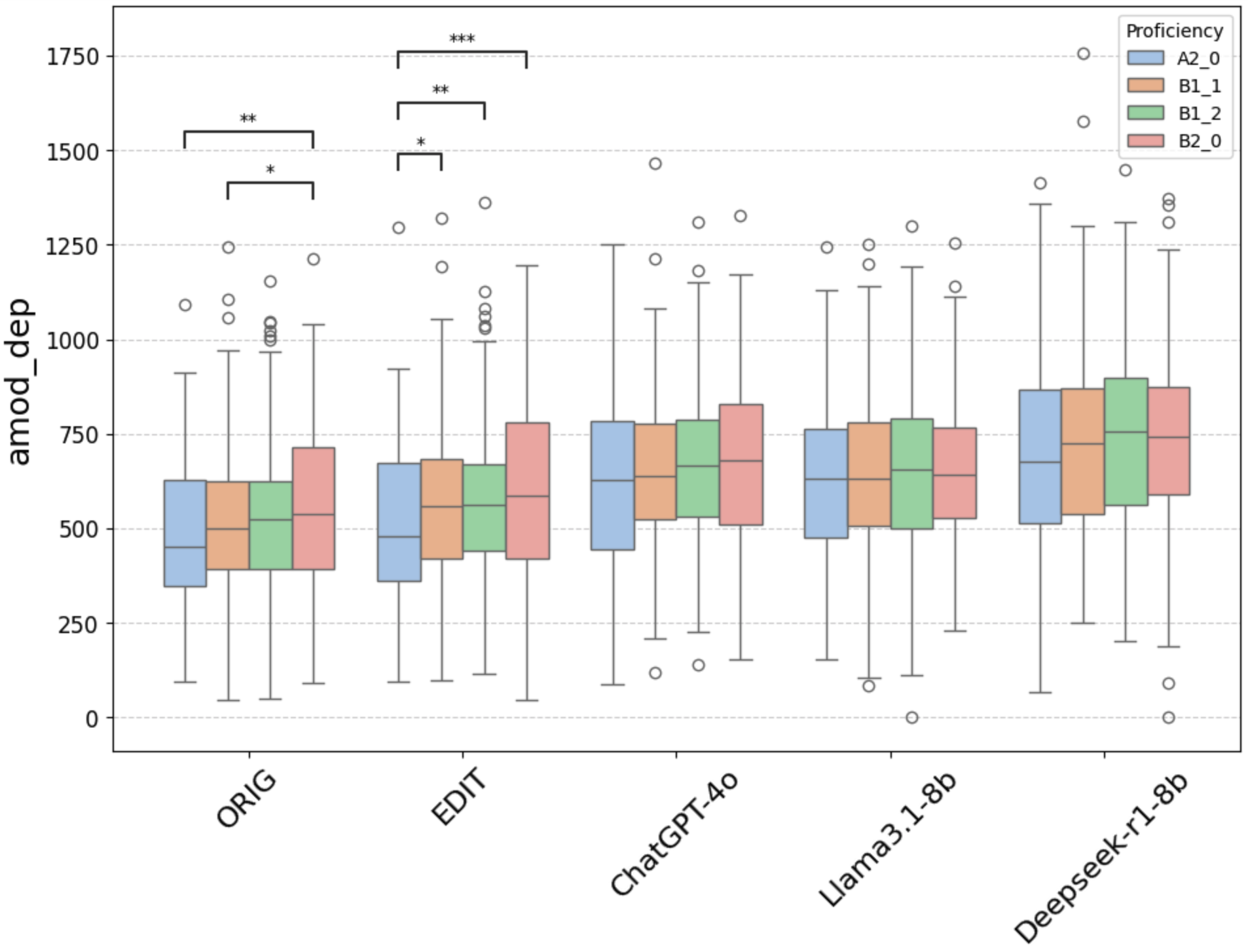}
    \caption{\texttt{amod\_dep} compared across ORIG, EDIT, and LLM-proofread texts by proficiency}
    \label{fig:4}
\end{figure}

At the phrase level, LLM proofreading increased the number of noun phrases (\texttt{np}), along with a rise in noun phrase dependencies (\texttt{np\_deps}). This suggests that LLM proofreading not only added more noun phrases but also enriched their internal structure. In particular, the marked increase in adjective modifier dependencies (\texttt{amod\_dep}; e.g., ``\underline{various} jobs'') suggests that LLM outputs favor more descriptive noun phrases (Figure \ref{fig:4}).

\begin{figure}[t]
    \centering
    \includegraphics[width=1\linewidth]{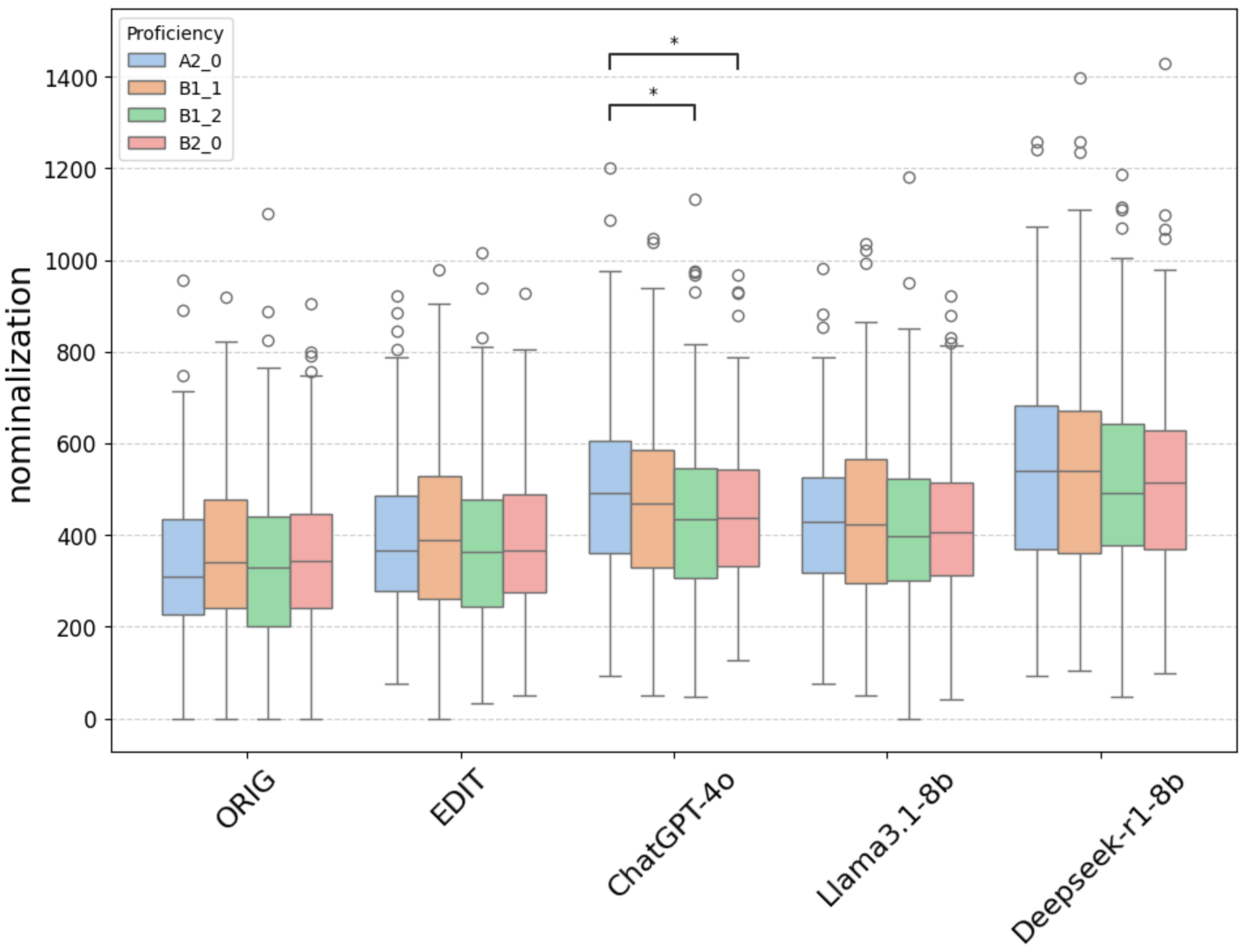}
    \caption{\texttt{nominalization} compared across ORIG, EDIT, and LLM-proofread texts by proficiency}
    \label{fig:5}
\end{figure}

At the morphological-syntactic level, both human and LLM  proofreading showed significant increases in \texttt{nominalization}, but the increases were more pronounced in the LLM outputs (Figure \ref{fig:5}). In contrast, the non-auxiliary use of the main verb ``be'' declined significantly under LLM proofreading, while human proofreading showed only a slight increase (\texttt{be\_mv}). Additionally, all proofreading modes consistently reduced the use of past tense (\texttt{past\_tense}).

\subsection{Cross-model consistency}

Based on the features that demonstrated meaningful group differences—and after removing indices with multicollinearity and conceptual overlap—we selected ten lexical or syntactic features. The composite lexical and syntactic scores exhibit strong internal consistency across the LLMs, with Cronbach's alpha values of 0.83 and 0.81, respectively. 

\begin{table}[!] 
\centering 
\resizebox{\linewidth}{!}{ 
\begin{tabular}{lcc} \toprule 
\textbf{Pair} & \textbf{Lexical} & \textbf{Syntax} \\ \midrule 
ChatGPT-4o – Llama3.1-8b & 0.70 & 0.62 \\ ChatGPT-4o – Deepseek-r1-8b & 0.60 & 0.53 \\ Llama3.1-8b – Deepseek-r1-8b & 0.56 & 0.65 \\ \bottomrule \end{tabular}} 
\caption{Pairwise Pearson correlations for lexical and syntactic features across LLMs} 
\label{tab:combined} \end{table}

Table \ref{tab:combined} presents the pairwise Pearson correlations among the three LLM proofreading models. For lexical features, ChatGPT-4o and Llama3.1-8b correlate at 0.70, while Deepseek-r1-8b correlates at 0.60 with ChatGPT-4o and 0.56 with Llama3.1-8b. For syntactic features, the corresponding correlations are 0.62, 0.53, and 0.65. These findings suggest that, despite minor variations, particularly with Deepseek-r1-8b, the LLMs tended to modify vocabulary and syntactic structures in a relatively consistent manner when proofreading L2 writings, as measured by our selected indices.

\section{Discussions}

We compared the lexical and syntactic features of original L2 writings with those of texts that were proofread by human and LLMs. We also evaluated the consistency of LLM proofreading across different models.

\paragraph{Lexical features} We found significant increases in bigram association strength, a ngram-level index of lexical sophistication, across all the proofreading modes. However, only LLM-proofread texts demonstrated notable changes in both word-level sophistication and diversity. Together, these results suggest that while both human and LLM proofreading improved the natural sequence of vocabulary--thus, enhancing the intelligibility of L2 writings--LLM proofreading provided an additional boost in lexical diversity and sophistication. In fact, this boost sometimes reduced or even eliminated typical differences between proficiency levels. Given that lexical sophistication and diversity are important constructs when evaluating L2 writing proficiency \cite{kyle2018tool,kyle2021assessing}, texts produced using LLM proofreading may obscure learners’ true writing abilities and artificially inflate their advanced language skills, ultimately undermining accurate assessment and long-term development.

We also observed that LLMs often replaced repeated words with alternative expressions—even when such changes are unwarranted—calling for caution. For example, ``I often \underline{can smell}'' became ``I often \underline{catch a whiff}”,  altering the intended meaning. Consequently, L2 writers using LLM proofreading should be mindful of unintended shifts in meaning or style and double-check suggested edits.

\paragraph{Syntactic features} 

Compared with the marked lexical shifts, syntactic edits were subtler but still distinct pattern of edits. First, both human and LLM proofreading consistently reduced past-tense verbs, favoring present or neutral tense---a pattern often associated with factual, persuasive prose \cite{burrough2003examining, fang2024time}. 

However, LLMs made more extensive structural modifications, including a higher proportion of non-finite clauses (e.g., ``Because the company that need worker will ask the job experiences'' $\rightarrow$ ``Companies \underline{looking} to \underline{hire} often require prior work experience'') and a marked increase in adjective modifier dependencies (e.g., ``become the social problem'' $\rightarrow$ ``become a \underline{significant} social problem''). They also introduced more nominalizations  (e.g., ``we should...'' $\rightarrow$ ``(our) primary \underline{responsibility}'') and reduced the non-auxiliary use of the main verb ``be'' (e.g., ``is not the first" $\rightarrow$ ``should not \underline{take} precedence'').

Meanwhile, although the increase in overall noun complexity following LLM proofreading was not statistically robust (\texttt{dp\_deps}), the gains were primarily driven by the insertion of adjective modifiers rather than by broader grammatical restructuring. For example, the structural complexity of noun phrases involving prepositional phrases (e.g., “disadvantages \underline{of works}”) or coordination (e.g., “advantages \underline{and} disadvantages”) remained largely unchanged.

\paragraph{Cross-model consistency} We found that the three LLMs exhibit generally consistent proofreading performance in terms of the major lexical and syntactic features. We speculate that this consistency arises from fundamental similarities in how they are trained and optimized for language generation tasks. Consequently, while different LLMs may produce distinct outputs, their overall patterns of lexical enhancement and syntactic restructuring remain comparable.

\section{Conclusions}

Our study shows that while both human and LLM proofreading improve lexical and syntactic features in L2 writing, LLMs typically implement more generative edits, reworking vocabulary and sentence structures to a greater extent. Although these changes may enhance clarity and style, they risk overshadowing the original meaning or authorial voice and potentially inflate apparent language proficiency. 

This finding has important implications for L2 writing practice. Acknowledging the great similarities in proofreading outcomes across different LLMs, more attention should be given to the question of ``how to use LLM-proofreading effectively'' rather than ``what LLM to use for proofreading.'' This key question can be addressed in reference to the observations that we have reported above, such as non-mandatory lexical substitution and excessive syntactic restructuring. Being aware of these tendencies in LLM-proofreading, L2 writers can better maintain control over their writing process while strategically making use of LLMs for linguistic improvements.

\section*{Limitations}

This study has several limitations. First, the same proofreading directive may be interpreted differently by human and LLM proofreaders, potentially affecting the nature and extent of the modifications. 

Second, the analysis lacks qualitative comparisons between original and edited texts, which could reveal subtler aspects of the revisions. As one reviewer noted, LLM-proofread essays may appear more sophisticated but sometimes sacrifice coherence or introduce unintended nuances, making them harder to read. A more systematic qualitative analysis (ideally supported by human perception data comparing human- and LLM-proofread texts) would clarify whether LLM edits genuinely improve writing quality or simply enhance surface-level features.

Third, the task effects and proficiency-level constraints limit generalizability: our analysis focused solely on argumentative writing by Asian university-level students who already possess a certain level of L2 English proficiency. Consequently, these findings may not extend to other types of writing or to L2 groups with different backgrounds.

%%%%%%%%%%%%%%%%%%%%%%%%%%%%%%%%

\clearpage
\newpage

\bibliography{custom}

\begin{thebibliography}{50}
\providecommand{\natexlab}[1]{#1}

\bibitem[{Achiam et~al.(2023)Achiam, Adler, Agarwal, Ahmad, Akkaya, Aleman, Almeida, Altenschmidt, Altman, Anadkat et~al.}]{achiam2023gpt}
Josh Achiam, Steven Adler, Sandhini Agarwal, Lama Ahmad, Ilge Akkaya, Florencia~Leoni Aleman, Diogo Almeida, Janko Altenschmidt, Sam Altman, Shyamal Anadkat, et~al. 2023.
\newblock \href {https://arxiv.org/abs/2303.08774} {Gpt-4 technical report}.
\newblock \emph{arXiv preprint arXiv:2303.08774}.

\bibitem[{Biber et~al.(2011)Biber, Gray, and Poonpon}]{biber2011should}
Douglas Biber, Bethany Gray, and Kornwipa Poonpon. 2011.
\newblock \href {https://doi.org/10.5054/tq.2011.244483} {Should we use characteristics of conversation to measure grammatical complexity in l2 writing development?}
\newblock \emph{Tesol Quarterly}, 45(1):5--35.

\bibitem[{Brysbaert et~al.(2014)Brysbaert, Warriner, and Kuperman}]{brysbaert2014concreteness}
Marc Brysbaert, Amy~Beth Warriner, and Victor Kuperman. 2014.
\newblock \href {https://link.springer.com/article/10.3758/s13428-013-0403-5} {Concreteness ratings for 40 thousand generally known english word lemmas}.
\newblock \emph{Behavior Research Methods}, 46:904--911.

\bibitem[{Bult{\'e} et~al.(2024)Bult{\'e}, Housen, and Pallotti}]{bulte2024complexity}
Bram Bult{\'e}, Alex Housen, and Gabriele Pallotti. 2024.
\newblock \href {https://onlinelibrary.wiley.com/doi/pdf/10.1111/lang.12669} {Complexity and difficulty in second language acquisition: A theoretical and methodological overview}.
\newblock \emph{Language Learning}.

\bibitem[{Bulté and Housen(2012)}]{bulte2012defining}
Bram Bulté and Alex Housen. 2012.
\newblock \href {https://benjamins.com/catalog/lllt.32} {Defining and operationalising l2 complexity}.
\newblock In Alex Housen, Folkert Kuiken, and Ineke Vedder, editors, \emph{Dimensions of L2 Performance and Proficiency: Complexity, Accuracy and Fluency in SLA}, pages 21--46. John Benjamins Publishing Company, Amsterdam.

\bibitem[{Burrough-Boenisch(2003)}]{burrough2003examining}
Joy Burrough-Boenisch. 2003.
\newblock \href {https://doi.org/10.1016/S0889-4906(01)00049-7} {Examining present tense conventions in scientific writing in the light of reader reactions to three dutch-authored discussions}.
\newblock \emph{English for Specific Purposes}, 22(1):5--24.

\bibitem[{Carduner(2007)}]{carduner2007teaching}
Jessie Carduner. 2007.
\newblock \href {https://doi.org/10.1080/09571730701317655} {Teaching proofreading skills as a means of reducing composition errors}.
\newblock \emph{Language Learning Journal}, 35(2):283--295.

\bibitem[{Coyne et~al.(2023)Coyne, Sakaguchi, Galvan-Sosa, Zock, and Inui}]{coyne2023analyzing}
Steven Coyne, Keisuke Sakaguchi, Diana Galvan-Sosa, Michael Zock, and Kentaro Inui. 2023.
\newblock \href {https://arxiv.org/abs/2303.14342} {Analyzing the performance of gpt-3.5 and gpt-4 in grammatical error correction}.
\newblock \emph{arXiv preprint arXiv:2303.14342}.

\bibitem[{Cronbach(1951)}]{cronbach1951coefficient}
Lee~J Cronbach. 1951.
\newblock Coefficient alpha and the internal structure of tests.
\newblock \emph{Psychometrika}, 16(3):297--334.

\bibitem[{Davis et~al.(2024)Davis, Caines, Andersen, Taslimipoor, Yannakoudakis, Yuan, Bryant, Rei, and Buttery}]{davis2024prompting}
Christopher Davis, Andrew Caines, O~Andersen, Shiva Taslimipoor, Helen Yannakoudakis, Zheng Yuan, Christopher Bryant, Marek Rei, and Paula Buttery. 2024.
\newblock \href {https://aclanthology.org/2024.findings-acl.711/} {Prompting open-source and commercial language models for grammatical error correction of english learner text}.
\newblock In \emph{Findings of the Association for Computational Linguistics ACL 2024}, pages 11952--11967.

\bibitem[{Fang and Maglio(2024)}]{fang2024time}
David Fang and Sam~J Maglio. 2024.
\newblock \href {https://www.sciencedirect.com/science/article/abs/pii/S0022103123001014} {Time perspective and helpfulness: Are communicators more persuasive in the past, present, or future tense?}
\newblock \emph{Journal of Experimental Social Psychology}, 110:104544.

\bibitem[{Fang et~al.(2023)Fang, Yang, Lan, Wong, Hu, Chao, and Zhang}]{fang2023chatgpt}
Tao Fang, Shu Yang, Kaixin Lan, Derek~F Wong, Jinpeng Hu, Lidia~S Chao, and Yue Zhang. 2023.
\newblock \href {https://arxiv.org/abs/2304.01746} {Is chatgpt a highly fluent grammatical error correction system? a comprehensive evaluation}.
\newblock \emph{arXiv preprint arXiv:2304.01746}.

\bibitem[{Guo et~al.(2025)Guo, Yang, Zhang, Song, Zhang, Xu, Zhu, Ma, Wang, Bi et~al.}]{guo2025deepseek}
Daya Guo, Dejian Yang, Haowei Zhang, Junxiao Song, Ruoyu Zhang, Runxin Xu, Qihao Zhu, Shirong Ma, Peiyi Wang, Xiao Bi, et~al. 2025.
\newblock \href {https://arxiv.org/pdf/2501.12948?} {Deepseek-r1: Incentivizing reasoning capability in llms via reinforcement learning}.
\newblock \emph{arXiv preprint arXiv:2501.12948}.

\bibitem[{Han et~al.(2024)Han, Yoo, Myung, Kim, Lim, Kim, Lee, Hong, Kim, Ahn, and Oh}]{han-etal-2024-llm}
Jieun Han, Haneul Yoo, Junho Myung, Minsun Kim, Hyunseung Lim, Yoonsu Kim, Tak~Yeon Lee, Hwajung Hong, Juho Kim, So-Yeon Ahn, and Alice Oh. 2024.
\newblock \href {https://doi.org/10.18653/v1/2024.customnlp4u-1.21} {{LLM}-as-a-tutor in {EFL} writing education: Focusing on evaluation of student-{LLM} interaction}.
\newblock In \emph{Proceedings of the 1st Workshop on Customizable NLP: Progress and Challenges in Customizing NLP for a Domain, Application, Group, or Individual (CustomNLP4U)}, pages 284--293, Miami, Florida, USA. Association for Computational Linguistics.

\bibitem[{Harwood(2018)}]{harwood2018proofreaders}
Nigel Harwood. 2018.
\newblock \href {https://journals.sagepub.com/doi/pdf/10.1177/0741088318786236} {What do proofreaders of student writing do to a master’s essay? differing interventions, worrying findings}.
\newblock \emph{Written Communication}, 35(4):474--530.

\bibitem[{Harwood et~al.(2009)Harwood, Austin, and Macaulay}]{harwood2009proofreading}
Nigel Harwood, Liz Austin, and Rowena Macaulay. 2009.
\newblock \href {https://www.sciencedirect.com/science/article/pii/S1060374309000216?casa_token=AhYGkO21x1kAAAAA:R0LTYB7t92ZpS-6ML67teWhxpSLqOw_DOnfTb2D5ILLhGEN4BZsWl5ihyOV-xwHdXRd_BL-tzA} {Proofreading in a uk university: Proofreaders’ beliefs, practices, and experiences}.
\newblock \emph{Journal of Second Language Writing}, 18(3):166--190.

\bibitem[{Heintz et~al.(2022)Heintz, Roh, and Lee}]{heintz2022comparing}
Kevin Heintz, Younghoon Roh, and Jonghwan Lee. 2022.
\newblock \href {https://escienceediting.org/journal/view.php?number=267&trk=organization_guest_main-feed-card_feed-article-content} {Comparing the accuracy and effectiveness of wordvice ai proofreader to two automated editing tools and human editors}.
\newblock \emph{Science Editing}, 9(1):37--45.

\bibitem[{Hyatt et~al.(2017)Hyatt, Bienenstock, and Tilan}]{hyatt2017student}
Jon-Philippe~K Hyatt, Elisa~Jayne Bienenstock, and Jason~U Tilan. 2017.
\newblock \href {https://doi.org/10.1152/advan.00109.2016} {A student guide to proofreading and writing in science}.
\newblock \emph{Advances in Physiology Education}, 41(3):324--331.

\bibitem[{Ishikawa(2018)}]{Ishikawa2018TheIE}
Shin'ichiro Ishikawa. 2018.
\newblock \href {https://api.semanticscholar.org/CorpusID:204784157} {The icnale edited essays: A dataset for analysis of l2 english learner essays based on a new integrative viewpoint}.
\newblock \emph{English Corpus Studies}, 25:117--130.

\bibitem[{Ishikawa(2021)}]{ishikawa2021asian}
Shin'ichiro Ishikawa. 2021.
\newblock \href {https://www.researchgate.net/profile/Shinichiro-Ishikawa-2/publication/364690608_Asian_Learners'_Knowledge_and_Use_of_L2_English_Words_and_Phrases_-A_Corpus-based_Study_on_Learners_in_China_Japan_Korea_and_Taiwan/links/63575dc06e0d367d91c4254a/Asian-Learners-Knowledge-and-Use-of-L2-English-Words-and-Phrases-A-Corpus-based-Study-on-Learners-in-China-Japan-Korea-and-Taiwan.pdf} {Asian learners’ knowledge and use of l2 english words and phrases: A corpus-based study on learners in china, japan, korea, and taiwan}.
\newblock In J.~Szerszunowicz, editor, \emph{Intercontinental Dialogue on Phraseology 4: Reproducible Language Units from an Interdisciplinary Perspective}, pages 493--510. University of Bialystok Publishing House.

\bibitem[{Jiang et~al.(2023)Jiang, Xu, Pan, He, and Xie}]{jiang2023exploring}
Zilu Jiang, Zexin Xu, Zilong Pan, Jingwen He, and Kui Xie. 2023.
\newblock \href {https://www.mdpi.com/2226-471X/8/4/247} {Exploring the role of artificial intelligence in facilitating assessment of writing performance in second language learning}.
\newblock \emph{Languages}, 8(4):247.

\bibitem[{Katinskaia and Yangarber(2024)}]{katinskaia2024gpt}
Anisia Katinskaia and Roman Yangarber. 2024.
\newblock \href {https://doi.org/10.5281/zenodo.1234567} {Gpt-3.5 for grammatical error correction}.

\bibitem[{Kyle et~al.(2018)Kyle, Crossley, and Berger}]{kyle2018tool}
Kristopher Kyle, Scott Crossley, and Cynthia Berger. 2018.
\newblock \href {https://link.springer.com/article/10.3758/s13428-017-0924-4} {The tool for the automatic analysis of lexical sophistication (taales): Version 2.0}.
\newblock \emph{Behavior Research Methods}, 50:1030--1046.

\bibitem[{Kyle and Crossley(2018)}]{kyle2018measuring}
Kristopher Kyle and Scott~A Crossley. 2018.
\newblock \href {https://onlinelibrary.wiley.com/doi/full/10.1111/modl.12468} {Measuring syntactic complexity in l2 writing using fine-grained clausal and phrasal indices}.
\newblock \emph{The Modern Language Journal}, 102(2):333--349.

\bibitem[{Kyle et~al.(2021)Kyle, Crossley, and Jarvis}]{kyle2021assessing}
Kristopher Kyle, Scott~A Crossley, and Scott Jarvis. 2021.
\newblock \href {https://www.tandfonline.com/doi/abs/10.1080/15434303.2020.1844205} {Assessing the validity of lexical diversity indices using direct judgements}.
\newblock \emph{Language Assessment Quarterly}, 18(2):154--170.

\bibitem[{Kyle and Eguchi(2021)}]{kyle2021automatically}
Kristopher Kyle and Masaki Eguchi. 2021.
\newblock \href {https://www.degruyterbrill.com/document/doi/10.21832/9781788924863-007/pdf?licenseType=restricted} {Automatically assessing lexical sophistication using word, bigram, and dependency indices}.
\newblock In Frances Blanchette and Constantine Lukyanenko, editors, \emph{Perspectives on the L2 Phrasicon: The View from Learner Corpora}, pages 126--151. De Gruyter Brill, Berlin.

\bibitem[{Kyle et~al.(2024)Kyle, Sung, Eguchi, and Zenker}]{kyle2024evaluating}
Kristopher Kyle, Hakyung Sung, Masaki Eguchi, and Fred Zenker. 2024.
\newblock \href {https://www.cambridge.org/core/journals/studies-in-second-language-acquisition/article/evaluating-evidence-for-the-reliability-and-validity-of-lexical-diversity-indices-in-l2-oral-task-responses/E51955F0291E21A916CD2C4787508B80} {Evaluating evidence for the reliability and validity of lexical diversity indices in l2 oral task responses}.
\newblock \emph{Studies in Second Language Acquisition}, 46(1):278--299.

\bibitem[{Laufer and Nation(1995)}]{laufer1995vocabulary}
Batia Laufer and Paul Nation. 1995.
\newblock \href {https://academic.oup.com/applij/article-abstract/16/3/307/184110} {Vocabulary size and use: Lexical richness in l2 written production}.
\newblock \emph{Applied Linguistics}, 16(3):307--322.

\bibitem[{Loem et~al.(2023)Loem, Kaneko, Takase, and Okazaki}]{loem2023exploring}
Mengsay Loem, Masahiro Kaneko, Sho Takase, and Naoaki Okazaki. 2023.
\newblock \href {https://aclanthology.org/2023.bea-1.18/} {Exploring effectiveness of gpt-3 in grammatical error correction: A study on performance and controllability in prompt-based methods}.
\newblock In \emph{Proceedings of the 18th Workshop on Innovative Use of NLP for Building Educational Applications (BEA 2023)}, pages 205--219.

\bibitem[{Lu(2010)}]{lu2010automatic}
Xiaofei Lu. 2010.
\newblock \href {https://www.jbe-platform.com/content/journals/10.1075/ijcl.15.4.02lu} {Automatic analysis of syntactic complexity in second language writing}.
\newblock \emph{International Journal of Corpus Linguistics}, 15(4):474--496.

\bibitem[{Lu(2011)}]{lu2011corpus}
Xiaofei Lu. 2011.
\newblock \href {https://doi.org/10.5054/tq.2011.240859} {A corpus-based evaluation of syntactic complexity measures as indices of college-level esl writers' language development}.
\newblock \emph{TESOL Quarterly}, 45(1):36--62.

\bibitem[{Meara and Bell(2001)}]{meara2001p}
Paul Meara and Huw Bell. 2001.
\newblock \href {https://eric.ed.gov/?id=EJ637430} {P-lex: A simple and effective way of describing the lexical characteristics of short l2 tests}.
\newblock \emph{Prospect}, 16(3):5--19.

\bibitem[{Meyer et~al.(2024)Meyer, Jansen, Schiller, Liebenow, Steinbach, Horbach, and Fleckenstein}]{meyer2024using}
Jennifer Meyer, Thorben Jansen, Ronja Schiller, Lucas~W Liebenow, Marlene Steinbach, Andrea Horbach, and Johanna Fleckenstein. 2024.
\newblock \href {https://www.sciencedirect.com/science/article/pii/S2666920X23000784} {Using llms to bring evidence-based feedback into the classroom: Ai-generated feedback increases secondary students’ text revision, motivation, and positive emotions}.
\newblock \emph{Computers and Education: Artificial Intelligence}, 6:100199.

\bibitem[{Nation and Beglar(2007)}]{nation2007vocabulary}
Paul Nation and David Beglar. 2007.
\newblock \href {https://openaccess.wgtn.ac.nz/articles/journal_contribution/A_vocabulary_size_test/12552197} {A vocabulary size test}.
\newblock \emph{The Language Teacher}, 31(7):9--13.

\bibitem[{Nelson et~al.(1998)Nelson, McEvoy, and Schreiber}]{nelson1998university}
Douglas~L Nelson, Cathy~L McEvoy, and Thomas~A Schreiber. 1998.
\newblock \href {https://link.springer.com/article/10.3758/bf03195588} {The university of south florida word association, rhyme, and word fragment norms}.
\newblock \emph{Behavior Research Methods, Instruments, \& Computers}, 36(3):402--407.

\bibitem[{Osawa(2024)}]{osawa2024integrating}
Koji Osawa. 2024.
\newblock \href {https://journals.sagepub.com/doi/full/10.1177/00336882231198913} {Integrating automated written corrective feedback into e-portfolios for second language writing: Notion and notion ai}.
\newblock \emph{RELC Journal}, 55(3):881--887.

\bibitem[{Salter-Dvorak(2019)}]{salter2019proofreading}
Hania Salter-Dvorak. 2019.
\newblock \href {https://www.sciencedirect.com/science/article/pii/S1475158518301577?casa_token=VtWKYMaYhmwAAAAA:wDclIhoY24iJluFkYCu1-zQyZsiTJ0i3sGgO_HDWri3K-Ub7R6hWjLtx0A4d3HkXzmAqQ4USOA} {Proofreading: How de facto language policies create social inequality for l2 master's students in uk universities}.
\newblock \emph{Journal of English for Academic Purposes}, 39:119--131.

\bibitem[{Sch{\"a}fer and Bildhauer(2012)}]{schafer2012}
Roland Sch{\"a}fer and Felix Bildhauer. 2012.
\newblock \href {https://aclanthology.org/L12-1497/} {Building large corpora from the web using a new efficient tool chain}.
\newblock In \emph{Proceedings of the Eighth International Conference on Language Resources and Evaluation (LREC'12)}, pages 486--493, Istanbul, Turkey. European Language Resources Association (ELRA).

\bibitem[{Shafto(2015)}]{shafto2015proofreading}
Meredith~A Shafto. 2015.
\newblock \href {https://www.mdpi.com/1660-4601/12/11/14445} {Proofreading in young and older adults: The effect of error category and comprehension difficulty}.
\newblock \emph{International Journal of Environmental Research and Public Health}, 12(11):14445--14460.

\bibitem[{Su et~al.(2023)Su, Lin, and Lai}]{su2023collaborating}
Yanfang Su, Yun Lin, and Chun Lai. 2023.
\newblock \href {https://doi.org/10.1016/j.asw.2023.100752} {Collaborating with chatgpt in argumentative writing classrooms}.
\newblock \emph{Assessing Writing}, 57:100752.

\bibitem[{Touvron et~al.(2023)Touvron, Lavril, Izacard, Martinet, Lachaux, Lacroix, Rozi{\`e}re, Goyal, Hambro, Azhar et~al.}]{touvron2023llama}
Hugo Touvron, Thibaut Lavril, Gautier Izacard, Xavier Martinet, Marie-Anne Lachaux, Timoth{\'e}e Lacroix, Baptiste Rozi{\`e}re, Naman Goyal, Eric Hambro, Faisal Azhar, et~al. 2023.
\newblock \href {https://arxiv.org/abs/2302.13971} {Llama: Open and efficient foundation language models}.
\newblock \emph{arXiv preprint arXiv:2302.13971}.

\bibitem[{Turner(2011)}]{turner2011rewriting}
Joan Turner. 2011.
\newblock \href {https://www.tandfonline.com/doi/abs/10.1080/03075071003671786} {Rewriting writing in higher education: The contested spaces of proofreading}.
\newblock \emph{Studies in Higher Education}, 36(4):427--440.

\bibitem[{Turner(2024)}]{turnerafterword}
Joan Turner. 2024.
\newblock \href {https://www.taylorfrancis.com/books/edit/10.4324/9781003334446/proofreading-editing-student-research-publication-contexts-nigel-harwood} {Afterword: Revisiting the boundaries of editing and proofreading}.
\newblock In Nigel Harwood, editor, \emph{Proofreading and Editing in Student and Research Publication Contexts: International Perspectives}, pages 221--233. Routledge, London.

\bibitem[{Warschauer et~al.(2023)Warschauer, Tseng, Yim, Webster, Jacob, Du, and Tate}]{warschauer2023affordances}
Mark Warschauer, Waverly Tseng, Soobin Yim, Thomas Webster, Sharin Jacob, Qian Du, and Tamara Tate. 2023.
\newblock \href {https://papers.ssrn.com/sol3/papers.cfm?abstract_id=4404380} {The affordances and contradictions of ai-generated text for writers of english as a second or foreign language}.
\newblock \emph{Journal of Second Language Writing}, 62:101071.

\bibitem[{Wilson et~al.(2014)Wilson, Olinghouse, and Andrada}]{wilson2014does}
Joshua Wilson, Natalie~G Olinghouse, and Gilbert~N Andrada. 2014.
\newblock \href {https://eric.ed.gov/?id=EJ1039856} {Does automated feedback improve writing quality?}
\newblock \emph{Learning Disabilities: A Contemporary Journal}, 12(1):93--118.

\bibitem[{Wu et~al.(2023)Wu, Wang, Wan, Jiao, and Lyu}]{wu2023chatgpt}
Haoran Wu, Wenxuan Wang, Yuxuan Wan, Wenxiang Jiao, and Michael Lyu. 2023.
\newblock \href {https://arxiv.org/abs/2303.13648} {Chatgpt or grammarly? evaluating chatgpt on grammatical error correction benchmark}.
\newblock \emph{arXiv preprint arXiv:2303.13648}.

\bibitem[{Xiao(2024)}]{xiao2024chatgpt}
Qimin Xiao. 2024.
\newblock \href {https://dl.acm.org/doi/abs/10.1145/3664934.3664946} {Chatgpt as an artificial intelligence (ai) writing assistant for efl learners: An exploratory study of its effects on english writing proficiency}.
\newblock In \emph{Proceedings of the 2024 9th International Conference on Information and Education Innovations}, pages 51--56.

\bibitem[{Yan and Zhang(2024)}]{yan2024l2}
Da~Yan and Shuxian Zhang. 2024.
\newblock \href {https://www.nature.com/articles/s41599-024-03543-y} {L2 writer engagement with automated written corrective feedback provided by chatgpt: A mixed-method multiple case study}.
\newblock \emph{Humanities and Social Sciences Communications}, 11(1):1--14.

\bibitem[{Zhao(2024)}]{zhao2024impact}
Dan Zhao. 2024.
\newblock \href {https://doi.org/10.1007/s10639-024-13145-5} {The impact of ai-enhanced natural language processing tools on writing proficiency: An analysis of language precision, content summarization, and creative writing facilitation}.
\newblock \emph{Education and Information Technologies}, 30(6):8055--8086.

\bibitem[{Zou and Huang(2024)}]{zou2024impact}
Min Zou and Liang Huang. 2024.
\newblock \href {https://link.springer.com/article/10.1007/s10639-023-12397-x} {The impact of chatgpt on l2 writing and expected responses: Voice from doctoral students}.
\newblock \emph{Education and Information Technologies}, 29(11):13201--13219.

\end{thebibliography}

\clearpage
\newpage

\appendix

\section{Descriptions of the selected indices}
\label{apeA}

\begin{table}[ht]
\begin{tabular}{lp{12cm}}
\toprule
\textbf{Index} & \textbf{Description} \\
\midrule
\multicolumn{2}{l}{Lexical indices} \\
\midrule
\texttt{ntypes} & Counts the number of unique words, taking into account their part-of-speech. \\
\texttt{mattr} & Computes the type-token ratio over a 50-word sliding window. \\
\texttt{b\_concreteness} & Uses psycholinguistic norms to assess word concreteness across categories based on large-scale ratings, indicating how tangible or abstract a word is perceived to be \cite{brysbaert2014concreteness}. \\
\texttt{usf} & Measures the number of distinct stimuli that elicit a target word in a word association experiment; lower USF scores suggest the use of words that are more contextually distinct \cite{nelson1998university}. \\
\texttt{cw\_lemma\_freq\_log} & Represents the logarithm of lemma frequencies for content words, computed with reference to an English web corpus \cite{schafer2012}. \\
\texttt{raw\_bg\_MI} & Calculates raw bigram mutual Information to quantify the strength of association between consecutive words, with higher values indicating a stronger collocational relationship; this is measured against an English web corpus. \\
\midrule

\multicolumn{2}{l}{Syntactic indices} \\
\midrule
\texttt{mltu} & Measures the average length of T-units, where a T-unit is defined as a main clause plus any subordinate clause(s) attached to it. \\
\texttt{all\_clauses} & Counts the total number of clauses in the text (normed by 10,000 words). \\
\texttt{nonfinite\_prop} & Computes the proportion of nonfinite clauses (e.g., gerunds, infinitives) relative to the total number of clauses. \\
\texttt{np} & Counts the total number of noun phrases, highlighting the nominal complexity within sentence structures (normed by 10,000 words). \\
\texttt{np\_deps} & Counts the number of internal dependencies within noun phrases (e.g., adjectives, prepositions, coordinations) (normed by 10,000 words). \\
\texttt{amod\_dep} & Measures the frequency of adjective modifier dependencies (normed by 10,000 words). \\
\texttt{nominalization} & Counts the frequency of nominalizations (i.e., words that convert verbs or adjectives into noun forms) identified by tokens containing predefined suffixes such as \textit{-al}, \textit{-ness}, among others (normed by 10,000 words). \\
\texttt{be\_mv} & Measures the frequency of the verb ``be'' when used as a main verb (excluding its auxiliary function) (normed by 10,000 words). \\
\texttt{past\_tense} & Measures the frequency of past tense verbs (normed by 10,000 words). \\
\bottomrule
\end{tabular}
\end{table}

\end{document}